\documentclass[twoside,a4paper,11pt]{article}
\usepackage{times}
\usepackage{elcvia}
\usepackage[english]{babel}
\usepackage{epsfig}
\usepackage{xcolor}
\usepackage{textcomp}
\usepackage{paralist}
\usepackage[inline]{enumitem}

\input{psfig.sty}

\newcommand{\vol}{16}            
\newcommand{\num}{2}            
\newcommand{\begpag}{\thepage}  
\newcommand{\finpag}{20}         
\newcommand{\yearp}{2017}       

\title{Learning audio and image representations with \\ bio-inspired trainable feature extractors
\footnote[0]{Correspondence to: Nicola Strisciuglio $<$n.strisciuglio@rug.nl$>$ \vspace{0.2cm}}
\footnote[0]{\noindent Recommended for acceptance by Anjan Dutta and Carles S\'anchez}
\footnote[0]{\noindent http://dx.doi.org/10.5565/rev/elcvia.1128} 
\footnote[0]{ELCVIA ISSN:1577-5097}
\footnote[0]{Published by Computer Vision Center / Universitat Aut\`onoma de 
Barcelona, Barcelona, Spain}
}

\author{Nicola Strisciuglio$^*$\\ 
\centerline{\small \em $^*$Johann Bernoulli Institute for Mathematics and Computer Science, University of Groningen, Netherlands} \\ \\
\centerline{\small \em This work was carried out at the University of Groningen (Netherlands) and at the University of Salerno (Italy)} \\ \\
\centerline{\small Received 24th Jul 2017; accepted 24th Nov 2017}
       }

\graphicspath{ {./figures/} }
\begin{document}

\maketitle

\pagestyle{myheadings}

\begin{abstract}
Recent advancements in pattern recognition and signal processing concern the automatic learning of data representations from labeled training samples. Typical approaches are based on deep learning and convolutional neural networks, which require large amount of labeled training samples. In this work, we propose novel feature extractors that can be used to learn the representation of single prototype samples in an automatic configuration process. We employ the proposed feature extractors in applications of audio and image processing, and show their effectiveness on benchmark data sets.
\end{abstract}

\markboth{\centerline{\small \it 
           Electronic Letters on Computer Vision and Image Analysis 
           \vol(\num ):\begpag-\finpag, \yearp}}
         {\centerline{\small \it 
           Electronic Letters on Computer Vision and Image Analysis
           \vol(\num):\begpag-\finpag, \yearp}}
\hrulefill



\section{Introduction}
Since when very young, we can quickly learn new concepts, and  distinguish between different kinds of object or sound. If we see a single object or hear a particular sound, we are then able to recognize such sample or even different versions of it in other scenarios. For example, if one sees an iron chair and associates  the object to the general concept of ``chairs'', he will be able to detect and recognize also wooden or wicker chairs. 
Similarly, when we hear the sound of a particular event, such as a scream, we are then able to recognize other kinds of scream that occur in different environments. We continuously learn representations of the real world, which we then use in order to understand new and changing environments. 

In the field of pattern recognition, traditional methods are typically based on representations of the real world that require a careful design of a suitable feature set (i.e. a data representation), which involves considerable domain knowledge and effort by experts. Recently, approaches for automated learning of representations from training data were introduced. Representation learning aims at avoiding engineering of hand-crafted features and providing automatically learned features suitable for the recognition tasks. Nowadays, widely popular approaches for representation learning are based on  deep learning techniques and convolutional neural networks (CNN). These techniques are very powerful, but are computationally expensive and require large amount of labeled training data to learn effective models for the applications at hand. 

In this paper we report the main achievements included in the doctoral thesis titled `Bio-inspired algorithms for pattern recognition in audio and image processing', in which we proposed novel approaches for \emph{representation learning} for audio and image signals~\cite{StrisciuglioThesis}.


\begin{figure}[!t]
\vspace{5mm}
\centering
\small
	\setlength{\unitlength}{42mm}
	\input{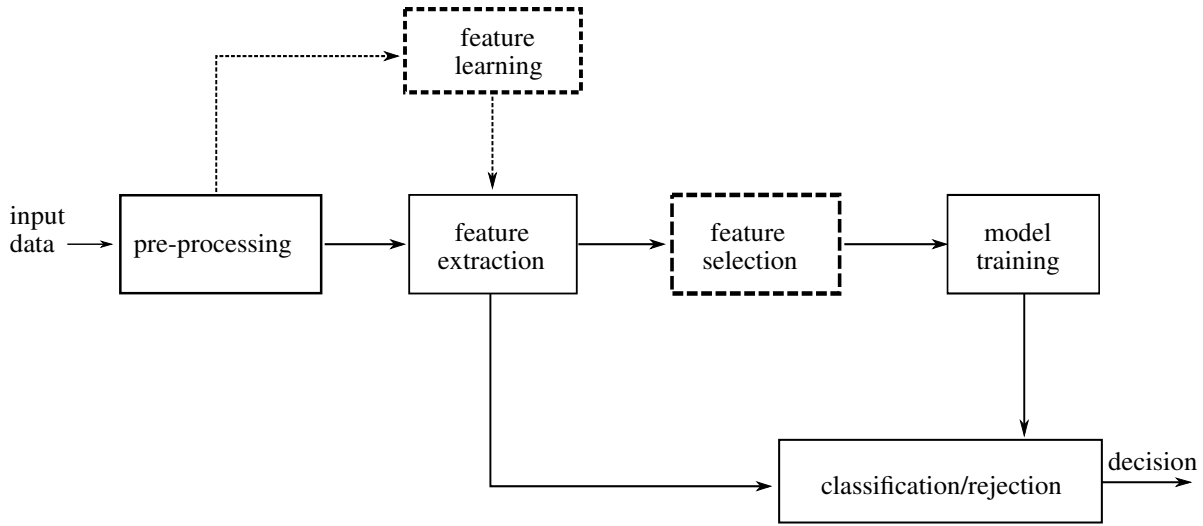}
	\caption{Overview of a pattern recognition system. The input data are pre-processed and then features are computed to extract important properties from such data. The features to be computed can be determined by an engineering process or can be learned from the data (\emph{representation learning}). Feature selection procedures can be employed to determine a sub set of discriminant features that are then used to train a classifier, which determines a model of the training data. Such model is then used in the operating phase of the system, while a classifier takes decisions on the input data.}
\label{fig:structure}
\end{figure}

\section{Motivation and contribution}
Motivated by the fact that we can learn effective representations of a new category of objects or sounds from a single example and successively generalize to a wide range of real-world samples, we studied the possibility of learning  data representations from small amounts of training samples. We investigated the design of feature extractors that can be automatically trained by showing single prototype samples, and employed them into pattern recognition systems to solve practical problems.

We proposed representation learning techniques for audio and image processing based on novel trainable feature extractors. The design and implementation of the proposed feature extractors are inspired by some functions of the human auditory and visual systems. The structure of the proposed feature extractors is learned from training samples in an automatic configuration step, rather than fixed a-priori in the implementation~\cite{StrisciuglioThesis}. We employed the newly designed methodologies into systems for audio event detection and classification in noisy environments and for delineation of blood vessels in retinal fundus images. The contributions of this work can be listed as:
\begin{inparaenum}[\itshape a\upshape)]
\item novel bio-inspired trainable feature extractors for representation learning of audio and image signals, respectively called COPE and \textit{B}-COSFIRE;
\item a system for audio event detection based on COPE feature extractors;
\item the release of two data sets of audio events of interest mixed to various background sounds and with different signal to noise ratios (SNR);
\item a method for delineation of elongated and curvilinear patterns in images based on \textit{B}-COSFIRE filters;
\item feature selection mechanisms based on information theory and machine learning approaches.
\end{inparaenum}

\section{Methods}
We introduced a novel approach for representation learning, based on trainable feature extractors. We extended the traditional scheme of pattern recognition systems  with feature learning algorithms (dashed box at the top of Figure~\ref{fig:structure}), which construct a suitable representation of training data by automatic configuring a set of feature extractors.

We proposed trainable COPE (Combination of Peaks of Energy) feature extractors for sound analysis, that can be trained to detect any sound pattern of interest. In an automatic configuration process performed on a single prototype sound pattern, the structure of a COPE feature extractor is learned by modeling the constellation of peak points in a  time-frequency representation of the input sound~\cite{CopePreliminary2015}. In the application phase, a COPE feature has high value when computed on the same sound used for configuration, but also to similar or corrupted versions of it due to noise or distortion. This accounts for generalization capabilities and robustness of detection of the patterns of interest. The response of a COPE feature extractor is computed as the combination of the weighted score of its constituent constellation of energy peaks. For further detail we refer the reader to~\cite{CopePreliminary2015}.
For the design of COPE feature extractors, we were inspired by some functions of the cochlea membrane and the inner hair cells in the auditory system, which convert the sound pressure waves into neural stimuli on the auditory nerve. 
We employed COPE feature extractors together with a multi-class Support Vector Machine (SVM) classifier to perform audio event detection and classification, also in cases where sounds have null or negative SNR. 

We proposed \textit{B}-COSFIRE (that stands for Bar-selective Combination of Shifted Filter Responses) filters for detection of elongated and curvilinear patterns in images and apply them to the delineation of blood vessels in retinal images~\cite{AzzopardiStrisciuglio,StrisciuglioVIP15}. The \emph{B}-COSFIRE filters are trainable, that is their structure is automatically configured from prototype elongated patterns.
The design of the \textit{B}-COSFIRE filters is inspired by the functions of some neurons, called simple cells,  in  area V1 of the visual system, which fire when presented with line or contour stimuli. A \textit{B}-COSFIRE filter achieves orientation selectivity by computing the weighted geometric mean of the output of a pool of Difference-of-Gaussians (DoG) filters, whose supports are aligned in a collinear manner. Rotation invariance is efficiently obtained by appropriate shiftings of the DoG filter responses. For further detail we refer the reader to~\cite{AzzopardiStrisciuglio}.

After configuring a large bank of \textit{B}-COSFIRE filters selective for vessels (i.e. lines) and vessel-endings (i.e. line-endings) of various thickness (i.e. scale), 
we proposed to use several approaches based on information theory and machine learning to select an optimal subset of \textit{B}-COSFIRE filters for the vessel delineation task~\cite{Strisciuglio15,Strisciuglio2016}. We indicate this procedure with the dashed box named `feature learning' in Figure~\ref{fig:structure}. We consider the selected filters as feature extractors to construct a pixel-wise feature vector which we used in combination with a SVM classifier to classify the pixels in the testing image as \emph{vessel} and \emph{non-vessel} pixels. 

\section{Experiments and Results}
We released two data sets for benchmark of audio event detection and classification methods, namely the MIVIA audio events~\cite{BoawPRL2015} and the MIVIA road events~\cite{ITS2015} data sets. We reported baseline results (recognition rate of  $86.7\%$ and $82\%$ on the two data sets) by using a real-time method for event detection that is based on an adaptation of the \emph{bag of features} classification scheme to noisy audio streams~\cite{BoawPRL2015,ITS2015}. The results that we achieved by using COPE feature extractors show a considerable improvement with respect to the ones of the bag of features approach. We obtained a recognition rate of $95.38\%$ on the MIVIA audio event and of $94\%$ (with standard deviation on cross-validation experiments equal to $4.32$) on the MIVIA road event data sets. We performed $t$-Student tests and observed a  statistically significant improvement of the recognition rate with respect to baseline performance on both data sets. 

We evaluated the performance of the proposed \textit{B}-COSFIRE filters on four  data sets of retinal fundus images for benchmarking of blood vessel segmentation algorithms, namely the DRIVE, STARE, CHASE\_DB1 and HRF data sets. 
The results that we achieved (DRIVE: Se = $0.7655$, Sp = $0.9704$; STARE: Se = $0.7716$, Sp~=~$0.9701$; CHASE\_DB1: Se = $0.7585$, Sp = $0.9587$; HRF: Se = $0.7511$, Sp = $0.9745$) are higher than the ones reported by many state-of-the-art methods based on filtering approaches. The filter selection procedure based on supervised learning that we proposed in~\cite{Strisciuglio2016} contributes to a statistically significant increase of performance results, which are higher than or comparable to the ones of other methods based on machine learning techniques.

We extended the application range of the \textit{B}-COSFIRE filters to aerial images for the delineation of roads and rivers, natural and textured images~\cite{StrisciuglioIWOBI17}, and to pavement and road surface images for the detection of cracks and damages~\cite{StrisciuglioCAIP17}. The results that we achieved are better than or comparable to the ones achieved by existing methods, which are usually designed to solve specific problems. The proposed \textit{B}-COSFIRE filters demonstrated to be effective in various applications and with different types of images (retinal fundus photography, aerial photography, laser scans) for delineation of elongated and curvilinear patterns.

We studied the computational requirements of the proposed algorithms in order to evaluate their applicability in real-world applications and the fulfillment of real-time constraints given by the considered problems. The MATLAB implementation of the proposed algorithms are publicly released for research purposes\footnote{The code is available from the GitLab repositories at  http://gitlab.com/nicstrisc}.

\section{Conclusions}
In this work, we proposed novel trainable feature extractors and employed them in applications of sound and image processing. The trainable character of the proposed feature extractors is in that their structure is learned directly from training data in an automatic configuration process, rather the fied in the implementation. This provides flexibility and adaptability of the proposed methods to different applications.  The experimental results that we achieved, compared to the ones of other existing approaches, demonstrated the effectiveness of the proposed methods in various applications.

This work contributes to the development of techniques for \emph{representation learning} in audio and image processing, suitable for domains where there is no availability of large amount of labeled training data.

\bibliographystyle{splncs03}
\bibliography{citations}

\end{document}